\documentclass[letterpaper, 10 pt, conference]{ieeeconf}  
\IEEEoverridecommandlockouts                              
\usepackage{graphicx}
\usepackage{nohyperref} 
\usepackage{url}  
\usepackage{flushend}
\usepackage{amsmath} 
\usepackage{amssymb}  
\usepackage{enumerate}
\usepackage{subfig}
\usepackage{colortbl}
\usepackage{booktabs}
\usepackage{cite}
\usepackage{fancyhdr}
\title{\LARGE \bf
On the Performance of ConvNet Features for Place Recognition 
}
\author{Niko S\"underhauf, Sareh Shirazi, Feras Dayoub, Ben Upcroft, and Michael Milford\\%
\thanks{The authors are with the ARC Centre of Excellence for Robotic Vision,
Queensland University of Technology (QUT). http://www.roboticvision.org/ email: {\tt\small
niko.suenderhauf@roboticvision.org}}%
}
\begin{document}
\maketitle
\thispagestyle{fancy}
\fancyhf{}
\fancyhead[OL]{\scriptsize Published in Proceedings of IEEE International Conference on Intelligent Robots and Systems (IROS), 2015. 
\\
\tiny \textcopyright 2015 IEEE. Personal use of this material is permitted. Permission from IEEE must be obtained for all other uses, in any current or future media, including reprinting/republishing this material for advertising or promotional purposes, creating new collective works, for resale or redistribution to servers or lists, or reuse of any copyrighted component of this work in other works.}
\addtolength{\headheight}{\baselineskip}
\begin{abstract}
After the incredible success of deep learning in the computer vision domain, 
there has been much interest in applying Convolutional Network (ConvNet) features in robotic fields 
such as visual navigation and SLAM. Unfortunately, there are fundamental differences 
and challenges involved. Computer vision datasets are very different in character 
to robotic camera data, real-time performance is essential, and performance priorities 
can be different. This paper comprehensively evaluates and compares the utility of 
three state-of-the-art ConvNets on the problems of particular relevance 
to navigation for robots; viewpoint-invariance and condition-invariance, and for the 
first time enables real-time place recognition performance using ConvNets with large maps 
by integrating a variety of existing (locality-sensitive hashing) and novel (semantic 
search space partitioning) optimization techniques. We present extensive experiments 
on four real world datasets cultivated to evaluate each of the specific challenges in 
place recognition. The results demonstrate that speed-ups of two orders of
magnitude can be achieved 
with minimal accuracy degradation, enabling real-time performance. We confirm that networks 
trained for semantic place \emph{categorization} also perform better at (specific) place 
\emph{recognition} when faced with severe appearance changes and provide a reference for which networks and layers are optimal for 
different aspects of the place recognition problem.
\end{abstract}
\section{Introduction}
\begin{figure}[t]
  \begin{center}
    \includegraphics[width=\linewidth]{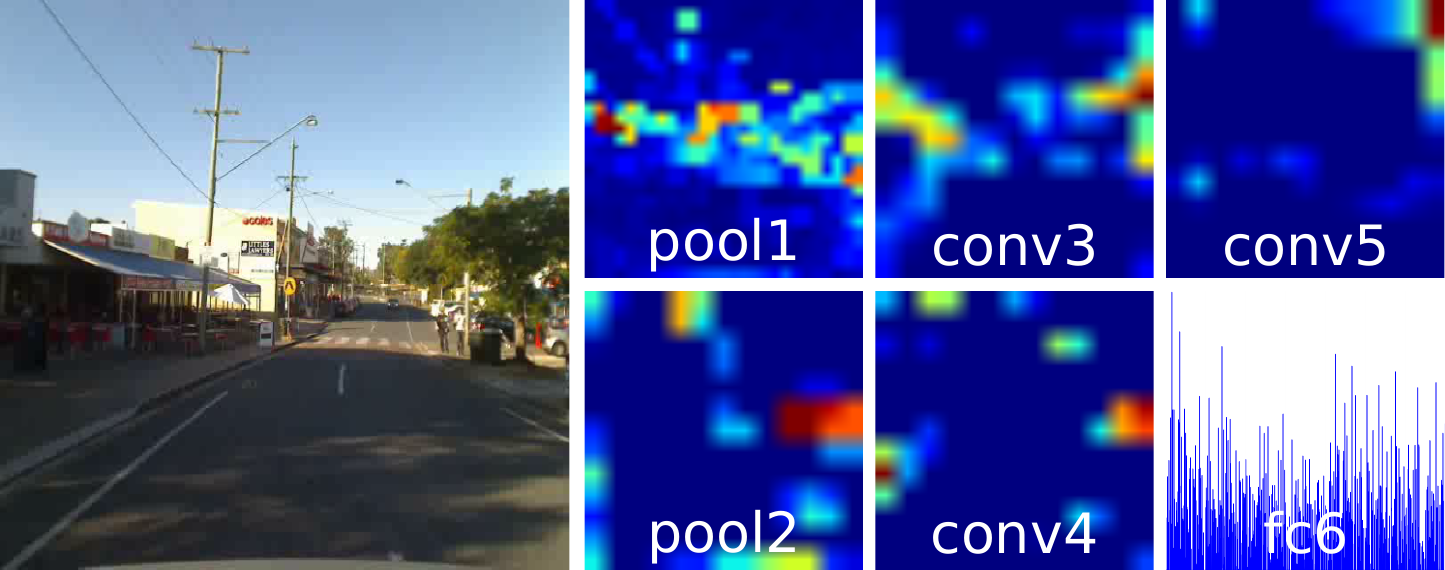}
  \end{center}
  \caption{
  Convolutional neural networks can extract features that serve as
  holistic image descriptors for place recognition. We found the features from layer \texttt{conv3}
  to be robust against appearance changes on a variety of datasets. The figure
  shows an example scene and extracted features from different network layers of
  the \texttt{AlexNet} network.
  }
  \label{fig:convNetFeatures}
\end{figure}

Robots that aim at autonomous long-term operations over extended periods of
time, such as days, weeks, or months, are faced with environments that can
undergo dramatic changes in their visual appearance over time. Visual place recognition --
the ability to recognize a known place in the environment using vision
as the main sensor modality -- 
is largely affected by these appearance changes 
and is therefore an active research field within the
robotics community. The recent literature proposes a variety of approaches to
address the challenges of this field \cite{Milford12,Corke13,Neubert14,Johns13,
Suenderhauf13b,Pepperell14,Naseer14,Upcroft14,Churchill12,McManus14a,Maddern14,Lowry14a,Stachniss14,Lowry14}.
Recent progress in the computer vision and machine learning community has shown
that the features generated by Convolutional Networks (ConvNets, see Fig. \ref{fig:convNetFeatures}
for examples) outperform other
methods in a broad variety of visual recognition, classification and detection
tasks \cite{Girshick14}. ConvNets have been demonstrated to be versatile and transferable,
i.e. even although they were trained on a very specific target task, they can be
successfully deployed for solving different problems and often even outperform
traditional hand engineered features~\cite{Razavian14}.
Our paper leverages these astounding properties and introduces the first real-time capable ConvNet-based place recognition
system. We exploit the hierarchical nature of ConvNet features and use the
semantic information encoded in the higher layers for search space partitioning and the
mid-level features for place matching under challenging conditions.
Locality-sensitive hashing of these features allows us to perform robust place
recognition against 100,000 known places with 3 Hz. 
We provide a thorough investigation of the utility of the
individual layers in the ConvNet hierarchy under severe appearance and viewpoint
variations and furthermore compare three state-of-the-art networks for the task
of place recognition.
We establish the following main results:
\begin{enumerate}
  \item features from the higher layers of the ConvNet hierarchy encode semantic information
    about a place and can be exploited to significantly reduce place
    recognition time by partitioning the search space,
  \item a speed-up of two orders of magnitude can be achieved by approximating
    the cosine distance between features with the Hamming distance over
    bitvectors obtained by Locality Sensitive Hashing,
    compressing the feature data by 99.6\% but retaining 95\% of place recognition
    performance;
    \item when comparing different ConvNets for the task of place
      \emph{recognition} under severe appearance changes, networks trained for the task of
    semantic place \emph{categorization} \cite{Zhou14}
    outperform the ConvNet trained for \emph{object} recognition;
  \item features from the middle layers in the ConvNet hierarchy exhibit
    robustness against appearance changes induced by the time of day,
    seasons, or weather conditions; and
  \item features extracted from the top layers are more robust with respect
    to viewpoint changes.
\end{enumerate}
In the following we review the related literature before describing the
datasets and evaluation protocol used. We analyze the performance of individual 
ConvNet feature layers for place recognition on several challenging datasets in
Section \ref{sec:experiments} and introduce important algorithmic performance
improvements in Section \ref{sec:real_time} before concluding the paper.
\section{Related Work}
\subsection{Place Recognition}
The focus of research in place recognition has recently moved from recognizing scenes without
significant appearance changes \cite{Cummins08, Cummins11, Sibley10} to more demanding, but also more realistic
changing environments.
Methods that address the place recognition problem span from matching sequences
of images \cite{Milford12, Johns13, Suenderhauf13b,Pepperell14,Naseer14}, transforming
images to become invariant against common scene changes such as shadows
\cite{Corke13, Upcroft14, McManus14a,Maddern14,Lowry14a}, learning how environments
change over time and predicting these changes in image space \cite{Neubert13b,
Lowry14a, Neubert14}, particle filter-based approaches that build up place recognition hypotheses
over time \cite{Maddern11, Stachniss14,Lowry14}, or build a map
of experiences that cover the different appearances of a place over time
\cite{Churchill12}.
Learning how the appearance of the environment changes generally requires
training data with known frame correspondences. \cite{Johns13} builds a database
of observed features over the course of a day and night. \cite{Neubert13b, Neubert14}
present an approach that learns systematic scene changes in order to improve
performance on a seasonal change dataset. \cite{McManus14} learns salient regions
in images of the same place with different environmental conditions. Beyond the
limitation of requiring training data, the generality of these methods is also
currently unknown; these methods have only been demonstrated to work in the same
environment and on the same or very similar types of environmental change to
that encountered in the training datasets.
\subsection{Convolutional Networks}
A commonality between all the aforementioned approaches is that they rely on a fixed set of
hand-crafted traditional features or operate on the raw pixel levels.
However, a recent trend in computer vision, and especially in the field
of object recognition and detection \cite{Girshick14,Russakovsky14}, is to exploit learned
features using deep convolutional networks (ConvNets).
It therefore appears very promising to analyze
these features and experimentally investigate their feasibility for the task of
place recognition.
Convolutional network is a well-known architecture and was proposed by
LeCun et al. in 1989 \cite{LeCun89} to recognize hand-written digits. 
Several research groups have recently shown that ConvNets outperform classical
approaches for object classification or detection that are based on hand-crafted
features \cite{Krizhevsky12, Sermanet13, Donahue13, Girshick14, Razavian14}.
The availability of pre-trained network models makes it easy to experiment with
such approaches for different tasks: The software packages \texttt{Overfeat}
\cite{Sermanet13} and \texttt{Caffe} \cite{Jia14} provide network
architectures pre-trained for a variety of recognition tasks.
\cite{Chen14} was the first to consider ConvNets for
place recognition. Our investigation is more thorough, since we cleanly separate
the performance contribution of the ConvNet features from the matching strategy, conduct more
systematic experiments on various datasets, compare three different ConvNets, and
contribute important algorithmic improvements which enable real-time performance. 
\begin{table}[b]
  \begin{center}
    \begin{tabular}{@{}llll@{}}	
      \toprule
      {\bf Layer} & {\bf Dimensions} & {\bf Layer} & {\bf Dimensions} \\ \midrule
      \texttt{pool1} & $96 \times 27 \times 27$ & \texttt{conv5} & $256 \times
      13 \times 13$  \\
      \texttt{pool2} & $256 \times 13 \times 13$ & \texttt{fc6} & $4096 \times 1 \times 1$  \\
      \texttt{conv3} & $384 \times 13 \times 13$ & \texttt{fc7} & $4096 \times 1
      \times 1$  \\
      \texttt{conv4} & $384 \times 13 \times 13$ &   \\ \bottomrule
    \end{tabular}
  \end{center}
  \caption{The layers from the \texttt{AlexNet} ConvNet used in our evaluation and
  their output dimensionality. 
  }
\label{tbl:layers}
\end{table}
\section{Preliminaries}
\subsection{Feature Extraction using a Convolutional Neural Network}
For the experiments described in the following section, we deploy the
\texttt{AlexNet} ConvNet \cite{Krizhevsky12} provided by \texttt{Caffe}
\cite{Jia14}. This network was pre-trained on the ImageNet
ILSVRC dataset \cite{Russakovsky14} for object recognition. It consists of five convolutional layers
followed by three fully connected layers and a soft-max layer.
The output of each individual layer can be extracted from the network and
used as a holistic image descriptor.
Since the third fully connected and the soft-max layer are adopted
specifically to the ILSVRC task (they have 1000 output neurons for the 1000
object classes in ILSVRC), we do not use them in the following experiments.
Table \ref{tbl:layers} lists the used layers and compares their dimensionality;
Fig. \ref{fig:convNetFeatures} displays some exemplar features extracted
from different layers. 
All images are resized to $231\times 231$ pixels to fit the expected input size of the ConvNet.
\subsection{Image Matching and Performance Measures}
Place recognition is performed by single-image nearest neighbor search based on the cosine
distance of the extracted feature vectors. 
We analyze the performance in terms of precision-recall curves and $F_1$ scores\footnote{
A match is considered a \emph{positive} if it passes a ratio test (ratio of the distances of the best over the second best match found in the nearest neighbor search) , and a \emph{negative} otherwise. Since every scene in the datasets has a ground truth match (i.e. there are no \emph{true negatives}), every \emph{negative} is a \emph{false negative}. A match is a \emph{true positive} when it is within $\pm 1..5$  frames of the ground truth (depending on the frame rate of the recorded dataset) and a \emph{false positive} otherwise.
The running parameter for creating the PR curves is the threshold on the ratio test.}.
In contrast to \cite{Chen14}
we explicitly do \emph{not} apply sequence search techniques or other
specialized algorithmic approaches to improve the matching performance in order
to observe the baseline performance of the ConvNet features under investigation.
\subsection{Datasets used in the Evaluation}
We used four different datasets in the experiments presented in Section
\ref{sec:experiments}. 
From the summary of their characteristics in Table \ref{tbl:datasets} we can see that
we carefully selected these datasets to form three groups of varying condition changes: Two datasets
exhibit severe appearance change but virtually no variation in viewpoints, one
shows viewpoint changes but only mild appearance changes, and three others
feature both types of variations. 
\begin{table}[b]
  \begin{center}
    \begin{tabular}{@{}llll@{}}	
      \toprule
       & & \multicolumn{2}{c}{\bf Variations in} \\
      {\bf Dataset} &  {\bf Environment}  & {\bf Appearance} & {\bf Viewpoint} \\ \midrule
      Nordland seasons & train journey &   severe & none   \\
      GP day-right--night-right & campus outdoor & severe & none   \\ \midrule
      GP day-left--night-right & campus outdoor & severe & medium  \\
      Campus Human--Robot & campus mixed & medium & medium  \\ 
      St. Lucia & suburban roads & medium & medium  \\ \midrule
      GP day-left--day-right & campus outdoor &  minor & medium  \\ \bottomrule
    \end{tabular}
  \end{center}
  \caption{The datasets used in the evaluation form three groups of varying
  severity of appearance and viewpoint changes.}
\label{tbl:datasets}
\end{table}
\subsubsection{The Nordland Dataset}
The Nordland dataset consists of video footage of a 10 hours long train journey
recorded from the perspective of the front cart in four different seasons.
Fig. \ref{fig:nordland:results} gives an impression of the
severe appearance changes between the different seasons. The Nordland dataset is
a perfect experimentation dataset since it exhibits no viewpoint variations
whatsoever and therefore allows to test algorithms on pure condition changes.
See \cite{Neubert14} for a more
elaborate discussion of this dataset. For our experiments we extracted image
frames at a rate of 1 fps and exclude all images that are taken inside tunnels
or when the train was stopped. We furthermore exclude images between the
timestamps 1:00h-1:30h and 2:47h-3:30h since these form a training dataset we
use in parallel work. 
\subsubsection{The Gardens Point Dataset}
The Gardens Point dataset has been recorded on the Gardens Point Campus of QUT
in Brisbane. It consists of three traverses of the environment, two during the
day and one during the night. The night images have been extremely contrast
enhanced and converted to grayscale in the process. One of the day traverses
was recorded keeping on the left side of the walkways, while the other day and
the night datasets have been recorded from the right side. This way, the dataset
exhibits both appearance and viewpoint changes.
\subsubsection{The Campus Human vs. Robot Dataset}
This dataset was recorded in different areas of our campus (outdoor, office,
corridor, food court) once by a robot using the Kinect RGB camera and once by a
human with a GoPro camera. While the robot footage was recorded during the day, the human traversed the environment during dawn, resulting in significant appearance changes especially in the outdoor scenes. The GoPro images were cropped to contain only the center part of half of the size of the original images. 
\subsubsection{The St. Lucia Dataset}
The St. Lucia dataset \cite{Glover10} has been recorded from inside a car moving
through a suburb in Brisbane at 5 different times during a day, and also on different days over a time
of two weeks. It features mild viewpoint variations due to slight changes in
the exact route taken by the car. More significant appearance changes due to
the different times of day can be observed, as well as some changes due to
dynamic objects such as traffic or cars parked on the street. 
\section{Layer-by-Layer Studies}
\label{sec:experiments}
This section provides a thorough investigation of the utility of different layers
in the ConvNet hierarchy for place recognition and evaluates their individual
robustness against the two main challenges in visual place recognition: severe
appearance changes and viewpoint variations.
\subsection{Appearance Change Robustness}
\begin{figure}[t]
  \begin{center}
       \includegraphics[width=0.4\linewidth]{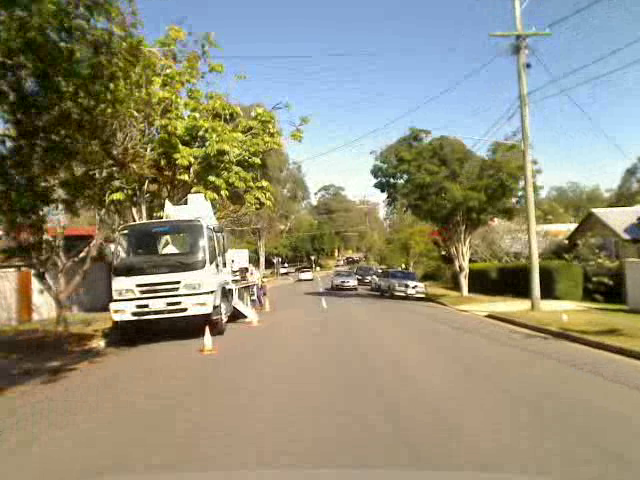}
       \includegraphics[width=0.4\linewidth]{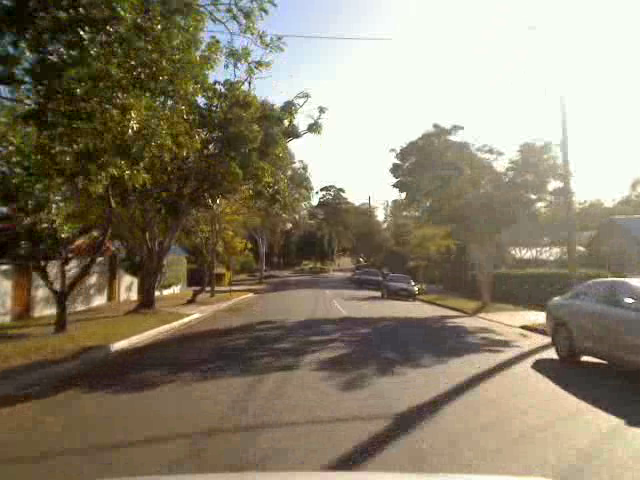}
       \includegraphics[width=0.9\linewidth]{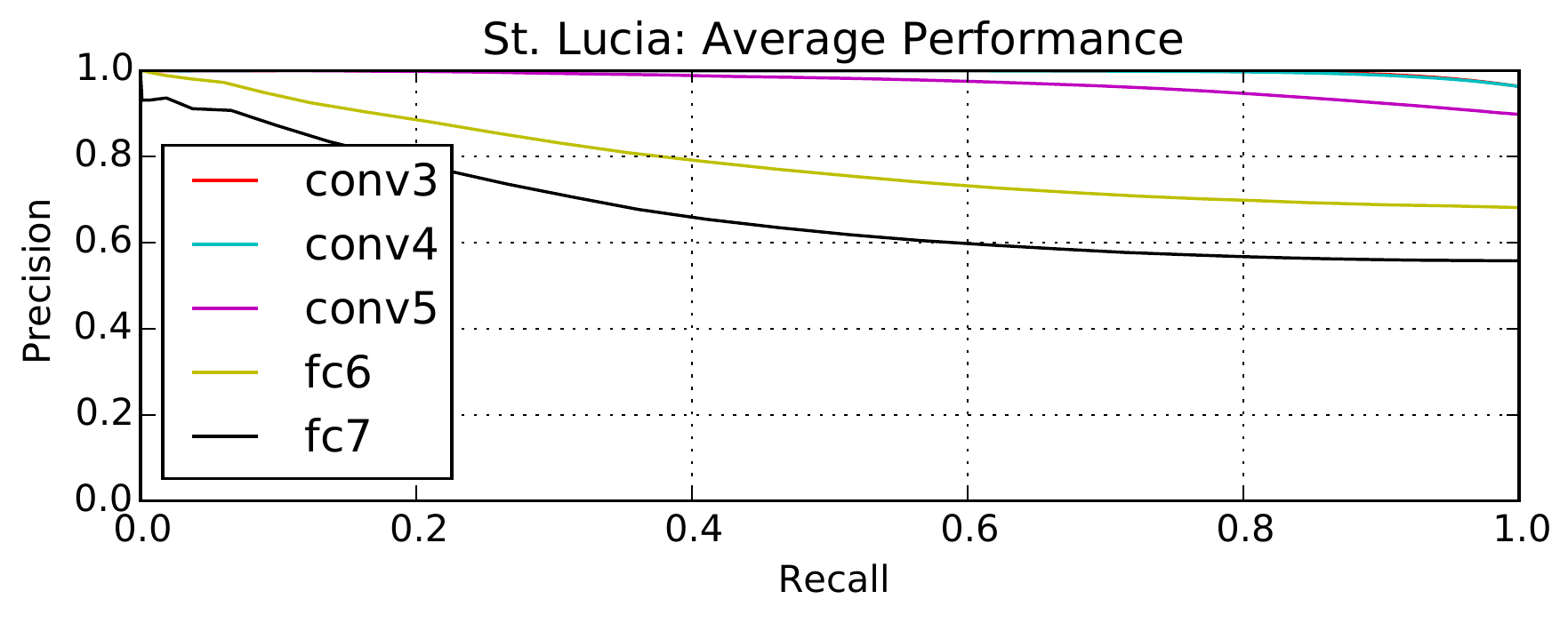}
  \end{center}
  \caption{Top: Two example images from the St. Lucia dataset showing the same
    place. 
    Bottom:
  Precision-recall curves averaged over all nine trials (recorded at different times
  during the day, and several days apart). Layers \texttt{conv3} and
  \texttt{conv4} perform almost perfectly (overlapping curves in the upper right
  corner). 
  }
  \label{fig:stLucia:results}
\end{figure}
\begin{figure}[t]
  \begin{center}
    \includegraphics[width=0.4\linewidth]{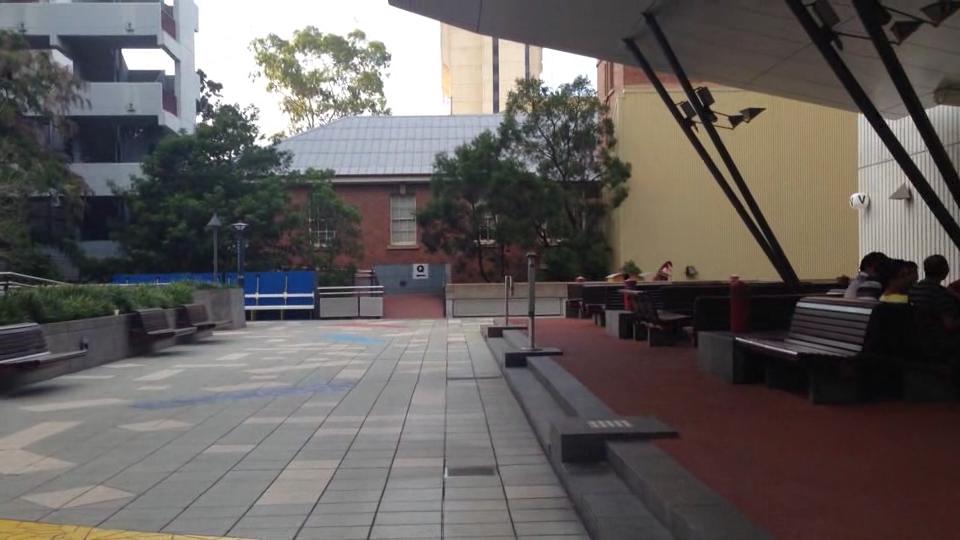}
    \includegraphics[width=0.4\linewidth]{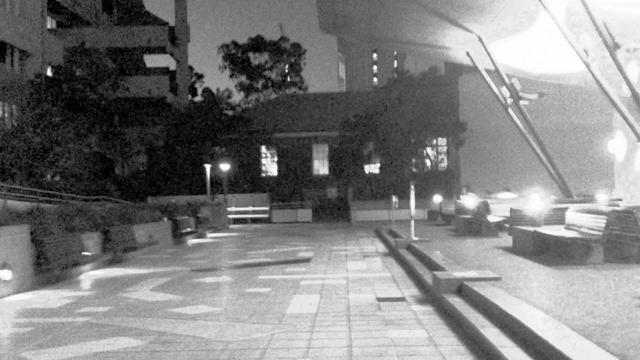}
    \includegraphics[width=0.9\linewidth]{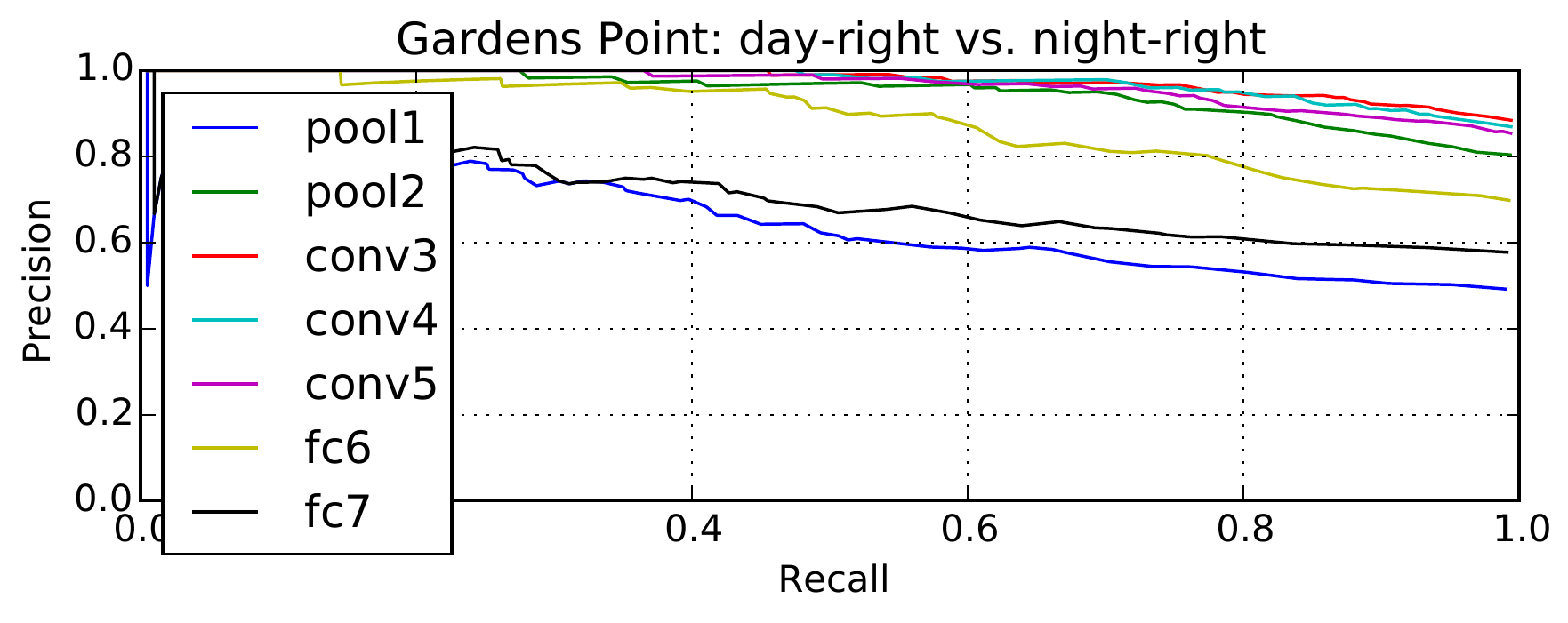}
  \end{center}
  \caption{Extreme appearance changes between day and night images. 
  The nighttime images have been contrast enhanced and converted to
  grayscale in the process. Despite these changes, layer \texttt{conv3} still performs reasonably well.}
  \label{fig:gardensPoint:day-right-night-right}
\end{figure}
\begin{figure}[t]
  \begin{center}
    \includegraphics[width=0.4\linewidth]{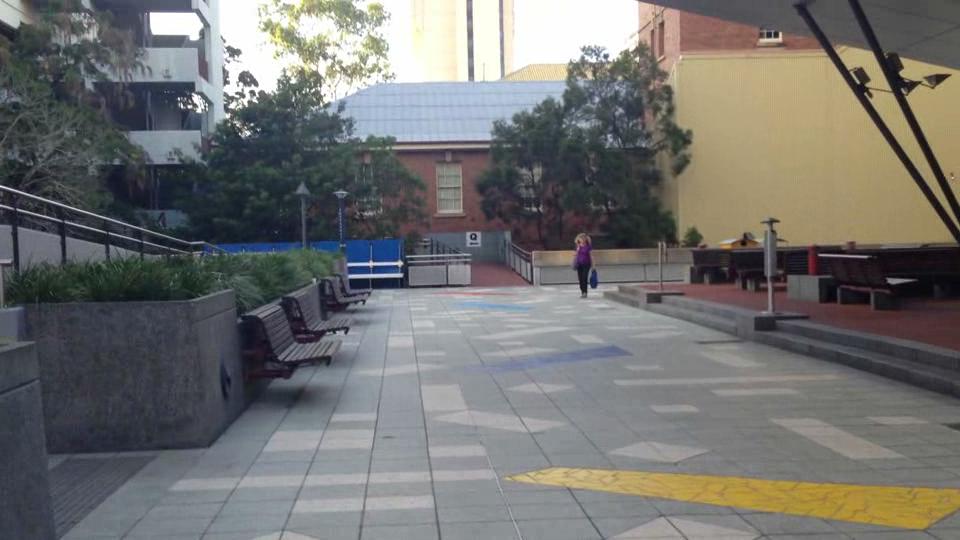}
    \includegraphics[width=0.4\linewidth]{figures/Image158002.jpg}
    \includegraphics[width=0.9\linewidth]{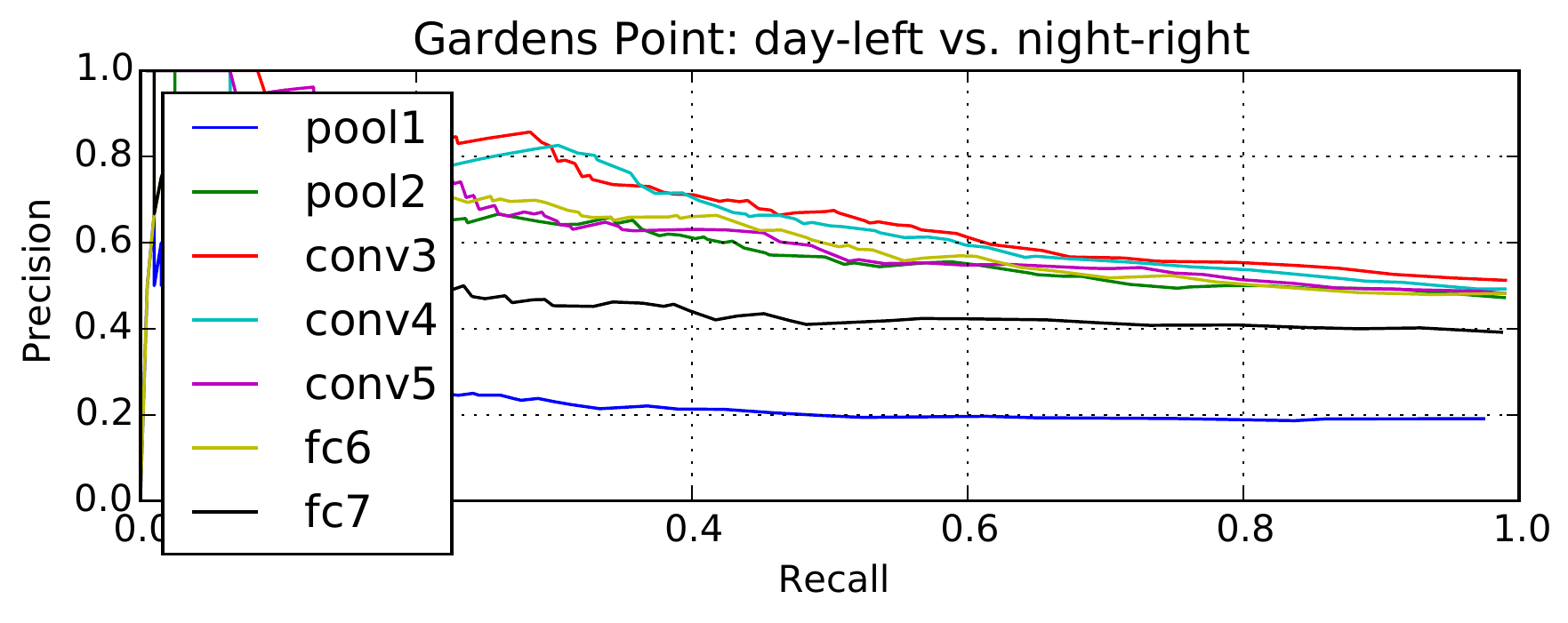}
  \end{center}
  \caption{When combining extreme appearance change with viewpoint changes,
  place recognition performance begins to deteriorate. This combination is still
  very challenging for current place recognition systems.
  }
  \label{fig:gardensPoint:day-left-night-right}
\end{figure}
\begin{figure}[t]
  \begin{center}
    \includegraphics[width=0.9\linewidth]{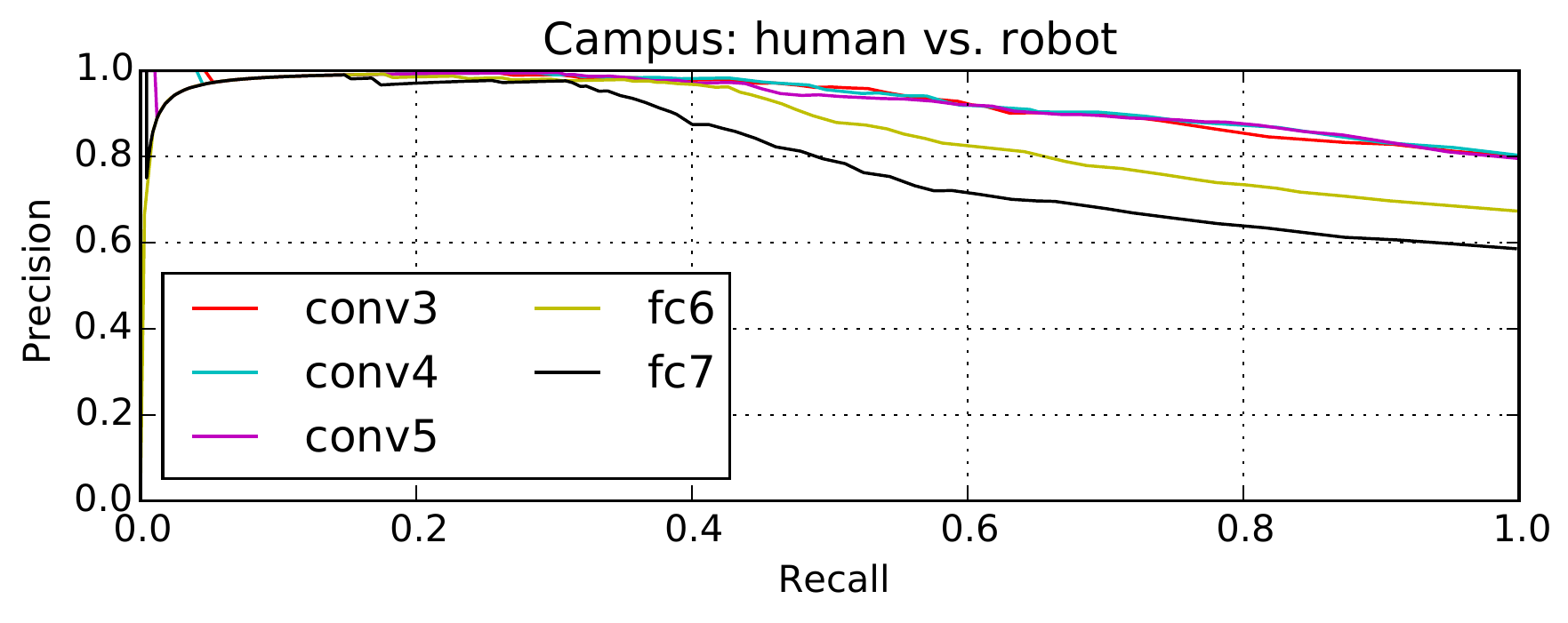}
   \end{center}
  \caption{The Campus dataset was recorded with different cameras by a robot during the day and a human at dawn.  
  The mid-level convolutional layers outperform the fully connected layers \texttt{fc6} and \texttt{fc7}.}
  \label{fig:campus}
\end{figure}
\begin{figure}[Th!B]
  \begin{center}
       \includegraphics[width=0.24\linewidth]{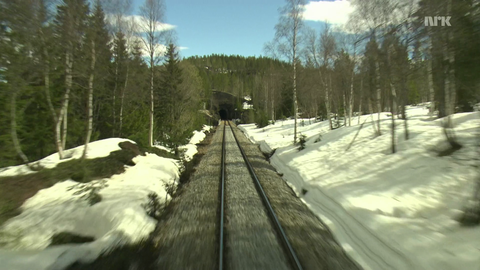}
    \includegraphics[width=0.24\linewidth]{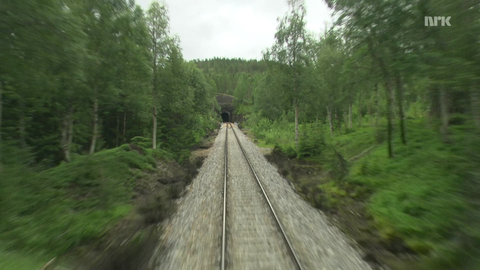}
    \includegraphics[width=0.24\linewidth]{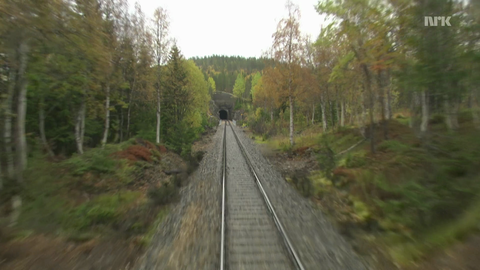}
    \includegraphics[width=0.24\linewidth]{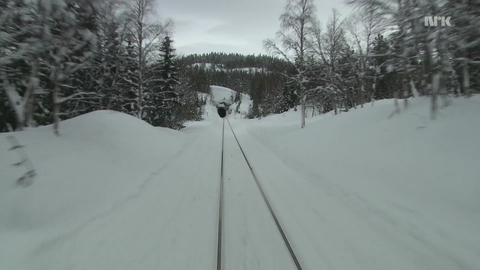}
       \includegraphics[width=0.9\linewidth]{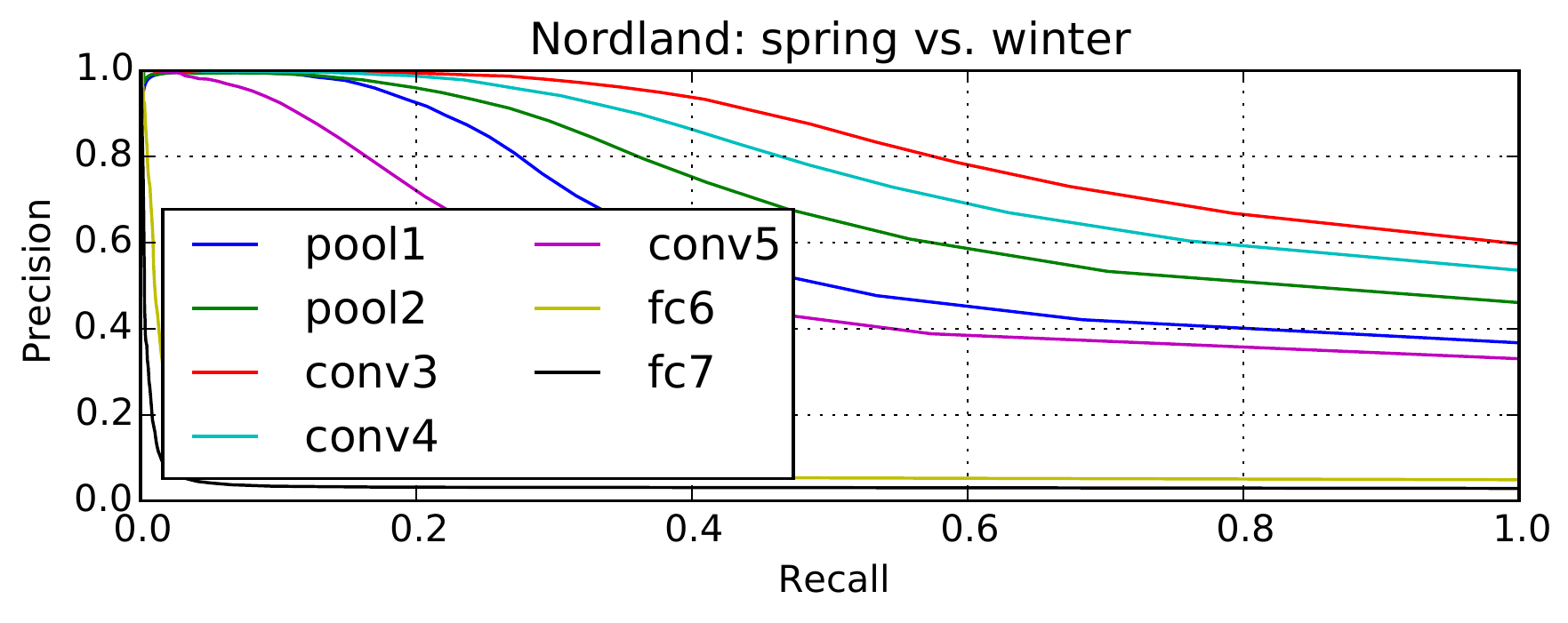}
       \includegraphics[width=0.9\linewidth]{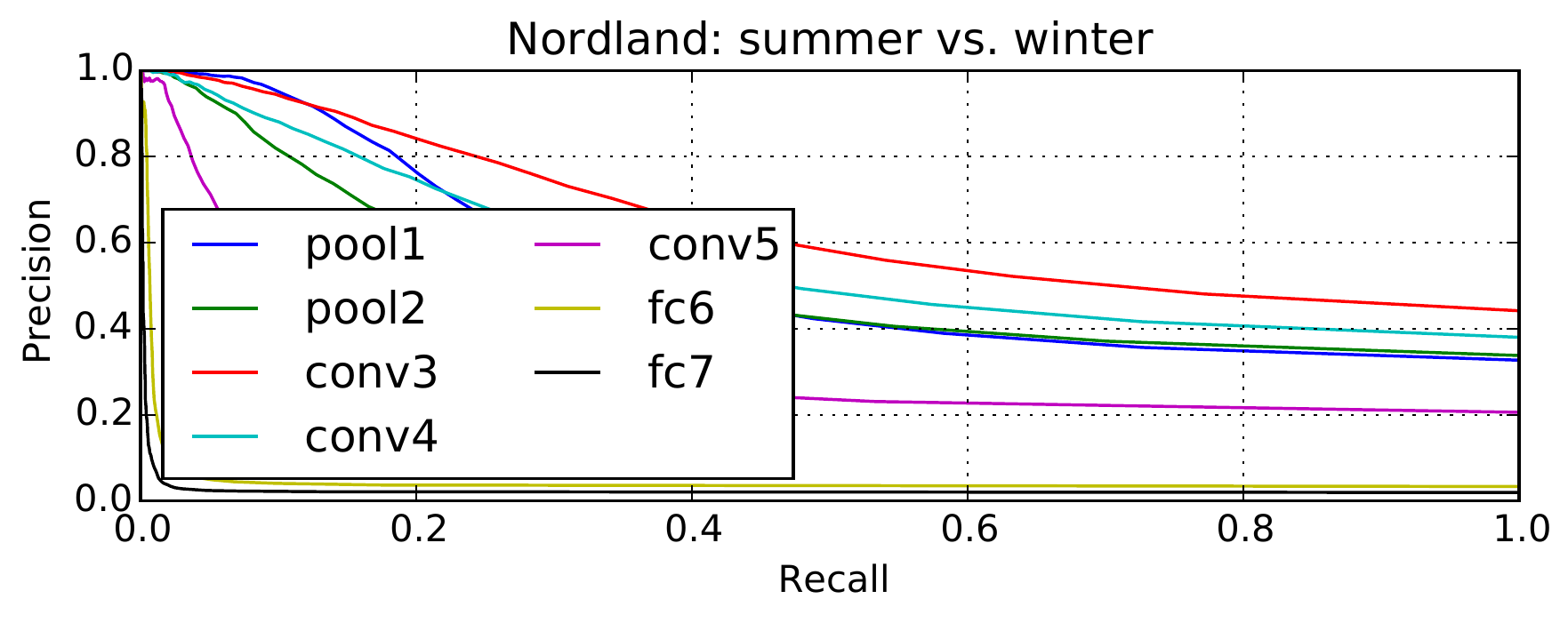}
       \includegraphics[width=0.9\linewidth]{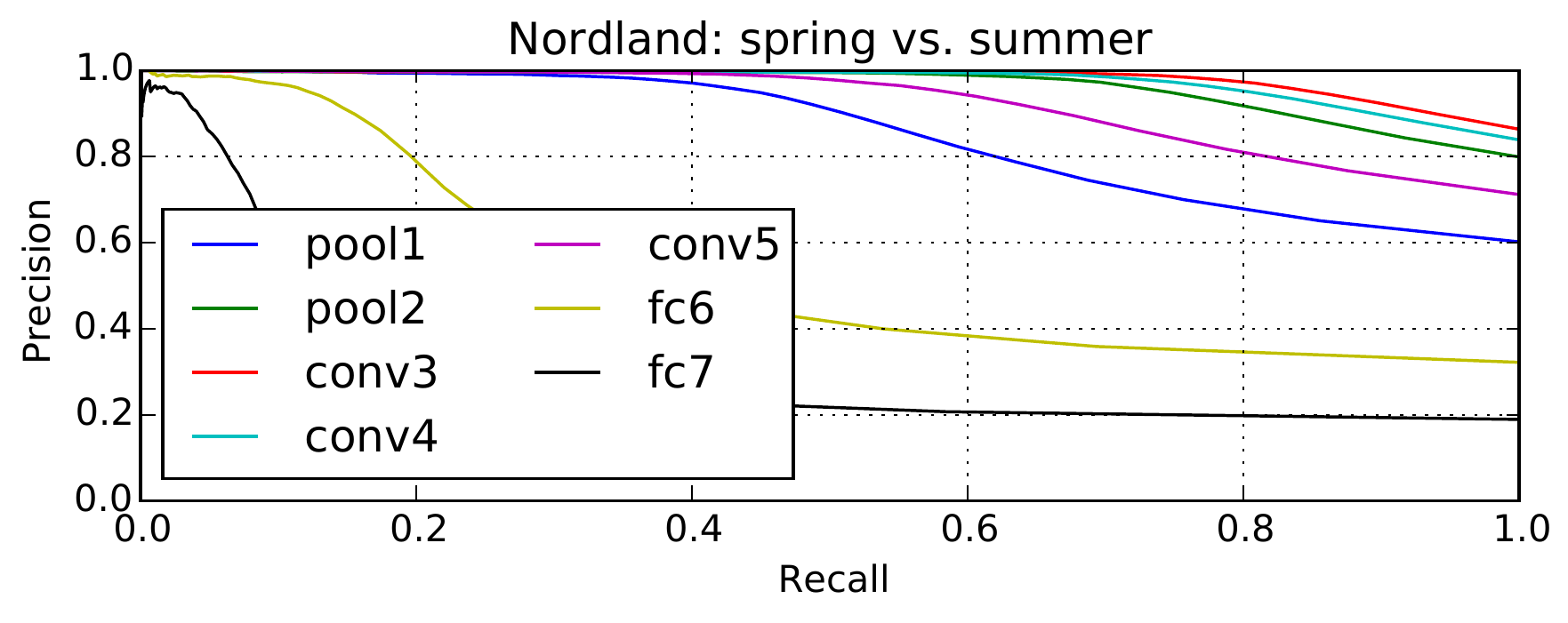}
  \end{center}
  \caption{Place recognition across seasons on the Nordland dataset.
  Despite the extreme appearance changes \texttt{conv3}
  performs acceptable, given that the image matching is based on a single frame
  nearest neighbor search only. Notice how \texttt{fc6} and \texttt{fc7} fail
  completely.
  }
  \label{fig:nordland:results}
\end{figure}
In a first set of experiments we analyze the robustness of the features from different layers in
the ConvNet architecture against appearance changes. We conduct these
experiments on the following datasets:
\begin{enumerate}
  \item the Nordland dataset, using 5 different season pairings that involve
    the spring or winter season
  \item the Gardens Point dataset, using all 3 subsets, spanning day and night 
  \item the St. Lucia dataset, using a total of 9 traversals from varying
    daytimes, over the course of several weeks
  \item the Campus dataset, using footage recorded by a human and a robot at
    different times of the day, including dawn
\end{enumerate}
Figures \ref{fig:stLucia:results}-\ref{fig:nordland:results} show the resulting
precision-recall curves for all experiments. Table \ref{tbl:results:F} summarizes the results
further and compares the $F_1$ scores with other state-of-the-art methods.
The experiments consistently show that the  mid-level features from layer
\texttt{conv3} are more robust against appearance changes than features from any
other layer.
Both the lower layers (\texttt{pool1}, \texttt{pool2}) and the higher layers
(e.g. \texttt{fc6} and \texttt{fc7}) in the feature hierarchy lack
robustness and exhibit inferior place recognition performance. 
We further compared our approach to SeqSLAM \cite{Milford12} a state-of-the-art method for place
recognition under extreme appearance changes and found that single image
matching using features extracted by layer \texttt{conv3}  matches or exceeds
SeqSLAM's performance. Previous work \cite{Neubert14,Glover08} established that
FAB-MAP \cite{Cummins08, Cummins11} is not capable of handling the severe
appearance changes of the Nordland and St. Lucia datasets.
\begin{table*}[bt]
  \begin{center}
    \begin{tabular}{@{}lllllllllll@{}}	
      \toprule
      & & \multicolumn{9}{c}{\bf $\bf{F_1}$-Scores} \\
       {\bf Dataset}  & & \multicolumn{7}{c}{\bf AlexNet Layers} & {\bf SeqSLAM} &{\bf FAB-MAP} \\
        & & \texttt{pool1} & \texttt{pool2} & \texttt{conv3} & \texttt{conv4} & \texttt{conv5} & \texttt{fc6} & \texttt{fc7} &  &\\ \midrule
       {\bf Nordland} & Spring -- Winter & 0.54 & 0.63 & 0.75 & 0.70 & 0.50 &
       0.09 & 0.06 & \hspace{0.3cm}{\bf 0.80} \textasteriskcentered & \hspace{0.4cm} $\dagger$ \\
       & Summer -- Winter & 0.49 & 0.51 & 0.61 & 0.55 & 0.34 & 0.07 & 0.04 &
       \hspace{0.3cm}{\bf 0.64} \textasteriskcentered & \hspace{0.4cm} $\dagger$ \\
       & Fall -- Winter & 0.52 & 0.60 & \bf 0.66 & 0.60 & 0.38 & 0.07 & 0.05 &
       \hspace{0.3cm}0.63 \textasteriskcentered  & \hspace{0.4cm} $\dagger$ \\
       & Spring -- Summer & 0.75 & 0.89 & \bf 0.93 & 0.91 & 0.83 & 0.49 & 0.32 & \hspace{0.3cm}0.86 \textasteriskcentered & \hspace{0.4cm} $\dagger$ \\
       & Spring -- Fall & 0.78 & 0.90 & \bf 0.93 & 0.92 & 0.84 & 0.53 & 0.35 & \hspace{0.3cm}0.88 \textasteriskcentered & \hspace{0.4cm} $\dagger$ \\
       \midrule
       {\bf Gardens Point} & day left -- day right & 0.61 & 0.88 & 0.89 & 0.88 &
       0.91 & \bf 0.96 & 0.96 & \hspace{0.3cm}0.44 & \hspace{0.3cm} ---   \\
       & day right -- night right & 0.66 & 0.89 & \bf 0.94 & 0.93 & 0.92 & 0.82 &
       0.73 & \hspace{0.3cm}0.44 & \hspace{0.3cm} ---  \\
       & day left -- night right & 0.32 & 0.64 & \bf 0.68& 0.66 & 0.65 & 0.65 &
       0.56 & \hspace{0.3cm}0.21 & \hspace{0.3cm}  --- \\
       \midrule
       {\bf St. Lucia} & average over 9 trials & --- & --- & \bf 0.98 & 0.98 & 0.95& 0.81& 0.72& \hspace{0.3cm} --- & \hspace{0.4cm} $\ddagger$ \\
       & worst trial & --- & --- &\bf 0.96 & 0.96 & 0.87 & 0.51 & 0.38 & \hspace{0.3cm}0.59 & \hspace{0.4cm} $\ddagger$ \\
       \midrule
       {\bf Campus} & human -- robot  & --- & --- & \bf 0.89 & 0.89 & 0.89 & 0.80 & 0.74 & \hspace{0.3cm}0.22 & \hspace{0.3cm} --- \\  
       \bottomrule
    \end{tabular}
  \end{center}
  \caption{Comparison of $F_1$-Scores between different layers and other
  state-of-the-art methods. $\dagger\,$We established in \cite{Neubert14} that FAB-MAP fails on the
  Nordland dataset. The maximum measured recall was 0.025 at 0.08 precision.
  $\ddagger\,$\cite{Glover08} reported that FAB-MAP is not able to
  address the challenges of the St. Lucia dataset, creating too many false
  positive matches.
  \textasteriskcentered For SeqSLAM on the Nordland dataset 
  we skipped every second frame (due to performance reasons) and used a sequence
  length of 5, resulting in an effective sequence length of 10. All other trials
  with SeqSLAM used a sequence length of 10.}
\label{tbl:results:F}
\end{table*}
\subsection{Viewpoint Change Robustness}
Viewpoint changes are a second major challenge for visual place recognition
systems. While point feature-based methods like FABMAP are less affected,
holistic methods such as the one proposed here are prone to error in the
presence of viewpoint changes. 
\begin{figure}[t]
  \begin{center}
    \includegraphics[width=0.2\linewidth]{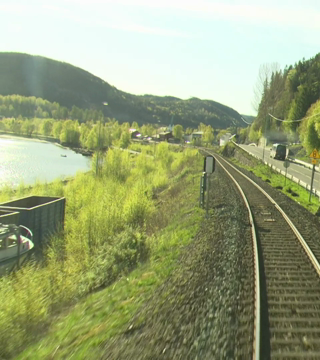}
    \hspace{0.5cm}
    \includegraphics[width=0.2\linewidth]{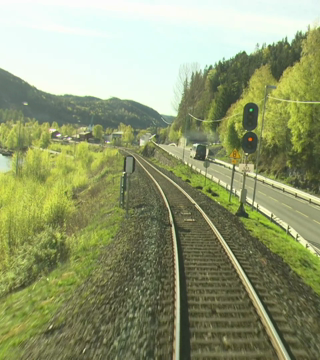}
    \includegraphics[width=0.9\linewidth]{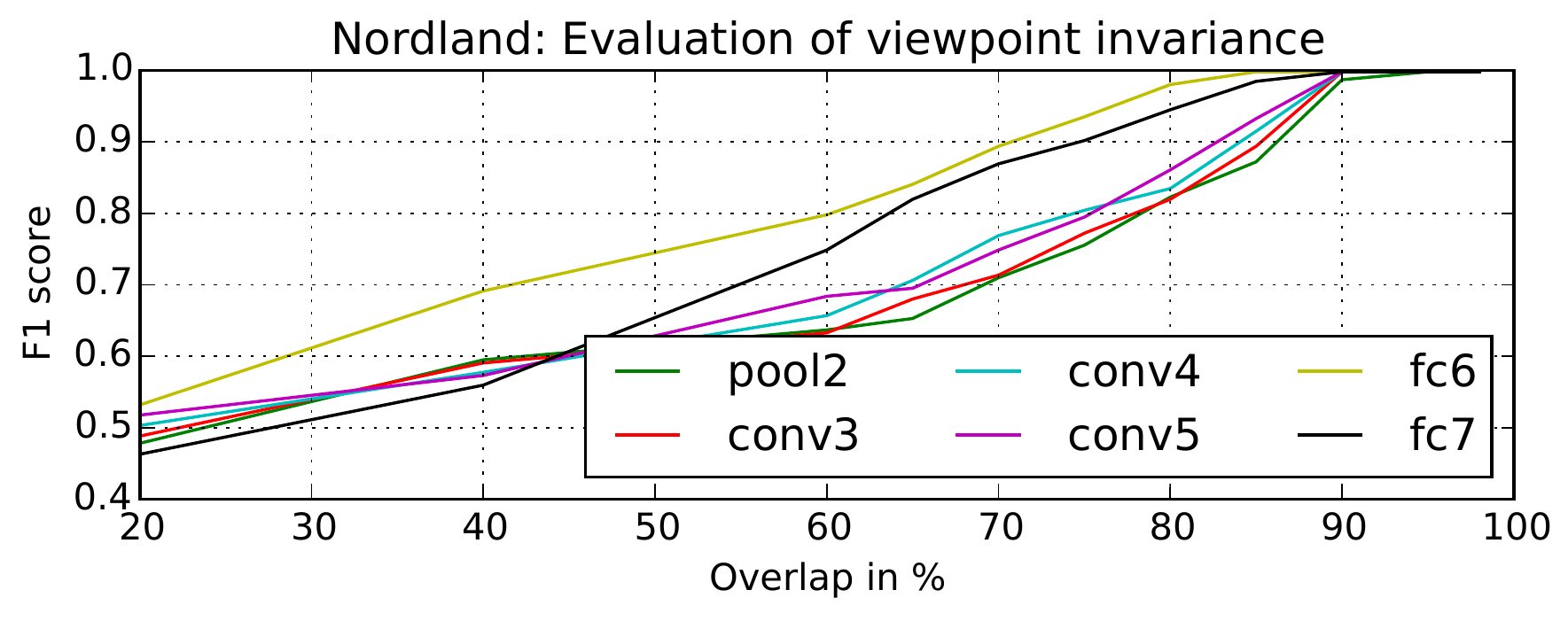}
  \end{center}
  \setlength{\belowcaptionskip}{-13pt}
  \caption{Top row: Examples for the synthetic viewpoint variation experiments conducted on the
  Nordland train dataset. We cropped the original images to half of their width
  and then created shifted versions with varying overlap. 
  Bottom: $F_1$ scores for different layers and overlap values. \texttt{fc6}
  performs best, but all layers are invariant to small variations that maintain 
  $\approx 90\%$ overlap.}
  \label{fig:nordland:translation:fscore}
\end{figure}
\begin{figure}[t]
  \begin{center}
    \includegraphics[width=0.4\linewidth]{figures/Image158.jpg}
    \includegraphics[width=0.4\linewidth]{figures/Image158001.jpg}
    \includegraphics[width=0.9\linewidth]{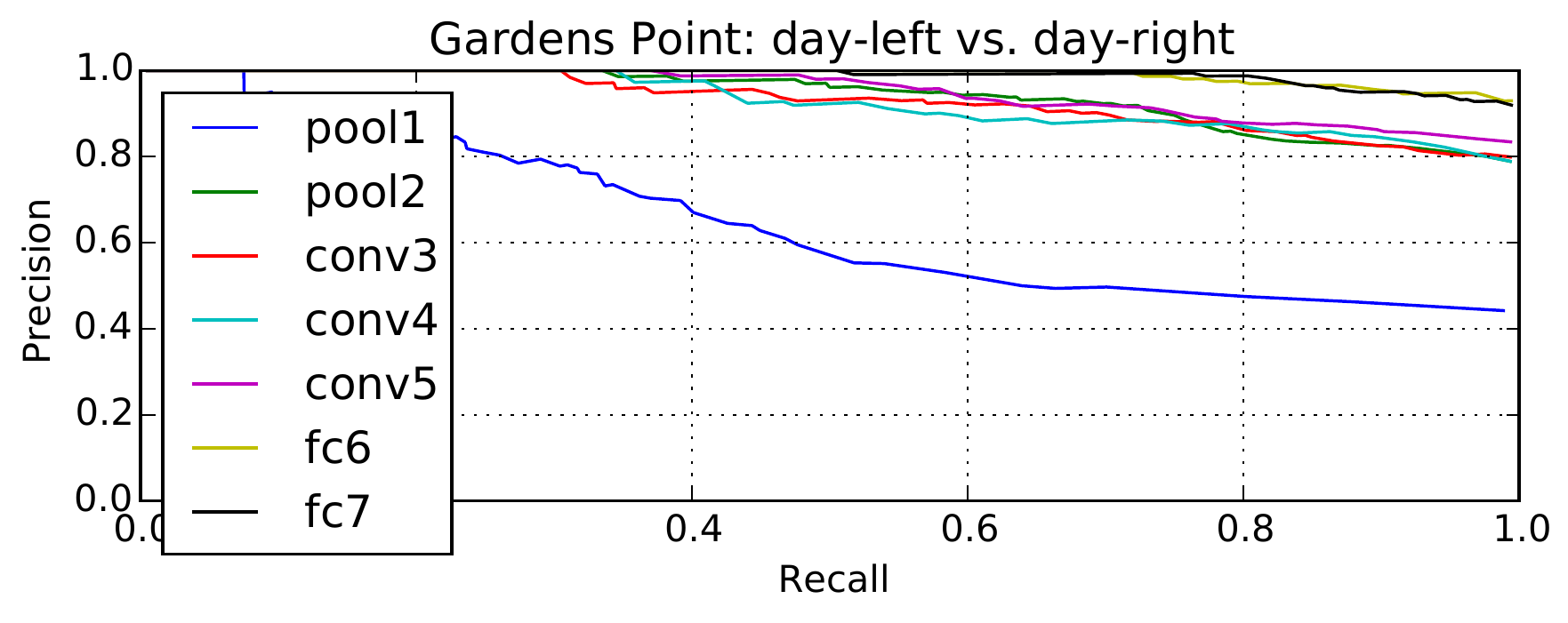}
  \end{center}
  \caption{Testing the viewpoint invariance on a realistic dataset. Two
  traverses through the same environment were recorded, one on the left and
  another on the right side of the walkway. The top two images compare the
  difference in viewpoint from the same place in both sequences. The plot
  shows that the higher layers \texttt{fc6} and \texttt{fc7} perform best,
  }
  \label{fig:gardensPoint:viewpoint}
\end{figure}
In order to quantify the viewpoint robustness of ConvNet features, we perform
two experiments:
\begin{enumerate}
   \item We use the Nordland spring dataset to create synthetic viewpoint changes
    by using shifted image crops.
  \item We use the Gardens Point dataset (day left vs. day right) to verify the
    observed effects on a real dataset.
\end{enumerate}
We conduct the first experiment on the Nordland spring dataset using images cropped to half
of the width of the original images.
We simulate viewpoint changes between two traverses by shifting the images of
the second traverse to the right. This results in overlaps between the images of
the simulated first and second traverse of between 98\% and 20\%.
For these experiments, the first 5000 images from the spring dataset (excluding
tunnels, stoppages and the training dataset mentioned before) were used.
Fig. \ref{fig:nordland:translation:fscore} illustrates
the results of the experiment with F-scores extracted from precision recall
statistics. 
In a second experiment we used the Gardens Point day-left vs. day-right dataset
that exhibits a lateral camera movement of 2-3 meters for medium viewpoint
changes between two traversals of the environment.
Only minor variations in the appearance of the scenes can be observed, mostly
caused by people walking on the campus.
Fig. \ref{fig:gardensPoint:viewpoint} shows the resulting precision recall
curves. 
Both experiments show that features from layers higher in the ConvNet
architecture, especially \texttt{fc6}, are more robust to viewpoint changes
than features from lower layers.
The increased viewpoint robustness of the upper layers can be accounted to the
\emph{pooling} layers that are part of the network architecture and perform
max-pooling after the first, second, and fifth convolutional layer.
The synthetic experiments summarized in Fig.
\ref{fig:nordland:translation:fscore} also show that all layers are
robust to mild viewpoint changes with more than 90\% overlap between scenes.
This effect is due to the resampling (resizing) of images before they are passed
through the network and the convolution and pooling operations that occur in the
first layer. In subsequent work \cite{Suenderhauf15a} we addressed the scenario of viewpoint and appearance changes occurring simultaneously using a number of new dataset.
\subsection{Summary and Discussion}
Table \ref{tbl:results} summarizes our experiments.
It is apparent that place recognition based on single image matching
performs well even under severe appearance changes when using the mid-level
\texttt{conv3} layer as a holistic feature extractor. Features from the higher
layers, especially \texttt{fc6}, only outperform \texttt{conv3} in situations
with viewpoint changes but none or only mild appearance variations.
This intuitively makes sense, since the features from the first convolutional
layers resemble very simple shape features \cite{Zeiler14} that are not discriminative
and generic enough to allow place recognition under severe appearance changes. 
The layers \emph{higher} in the hierarchy, and especially the fully connected layers,
are more semantically meaningful but therefore lose their ability to discriminate
between individual places within the same semantic type of scene. 
For example in the St. Lucia footage, the higher layers would encode the
information 'suburban road scene' equally for all images in the dataset. This
enables place \emph{categorization} but is disadvantageous for place \emph{recognition}.
The mid-level layers -- and especially \texttt{conv3} -- seem to encode just the right amount of information; they are more informative and more robust to changes than pure low-level pixel or gradient based features, while remaining discriminate enough to identify individual places.
\begin{table}[b]
  \begin{center}
    \begin{tabular}{@{}llll@{}}	
      \toprule
       & \multicolumn{2}{c}{\bf Variations in} & \\
      {\bf Dataset} & {\bf Appearance} & {\bf Viewpoint} & {\bf Best Layer}\\ \midrule
      Nordland seasons & severe & none & \texttt{conv3}  \\
      GP day-right vs night-right & severe & none & \texttt{conv3}  \\ \midrule
      GP day-left vs night-right & severe & medium & \texttt{conv3}  \\
      St. Lucia & medium & medium & \texttt{conv3} \\
      Campus & medium & medium & \texttt{conv3} \\ \midrule
      GP day-left vs day-right & minor & medium & \texttt{fc6} \\
      Nordland synthetic & none & \emph{varied} & \texttt{fc6} \\ \bottomrule
    \end{tabular}
  \end{center}
  \caption{The mid-level \texttt{conv3} outperforms all other layers in the presence of
  significant appearance changes.}
\label{tbl:results}
\end{table}
Extracting a ConvNet feature from an image requires approximately 15 ms on a
Nvidia Quadro K4000 GPU. The bottleneck of the place recognition system is the
nearest neighbor search that is based on the cosine distance between 64,896
dimensional feature vectors -- a computationally expensive operation. Our
Numpy/Scipy implementation requires 3.5 seconds to find a match among 10,000
previously visited places.
As we shall see in the next section, two important algorithmic improvements can be
introduced that lead to a speed-up of 2 orders of magnitude.
\section{Real-Time Large-Scale Place Recognition}
\label{sec:real_time}
In contrast to typical computer vision benchmarks where the recognition accuracy
is the most important performance metric, robotics applications depend on
agile algorithms that can provide a solution under certain soft
real-time constraints. 
The nearest neighbor search is the key limiting factor for large-scale place recognition, as its runtime
is proportional to the number of stored previously visited places.
In the following we will explore two approaches that will decrease the required search time by
two orders of magnitude with only minimal accuracy degradation.
\subsection{Locality Sensitive Hashing for Runtime Improvements}
\begin{figure}[t]
  \begin{center}
    \includegraphics[width=0.9\linewidth]{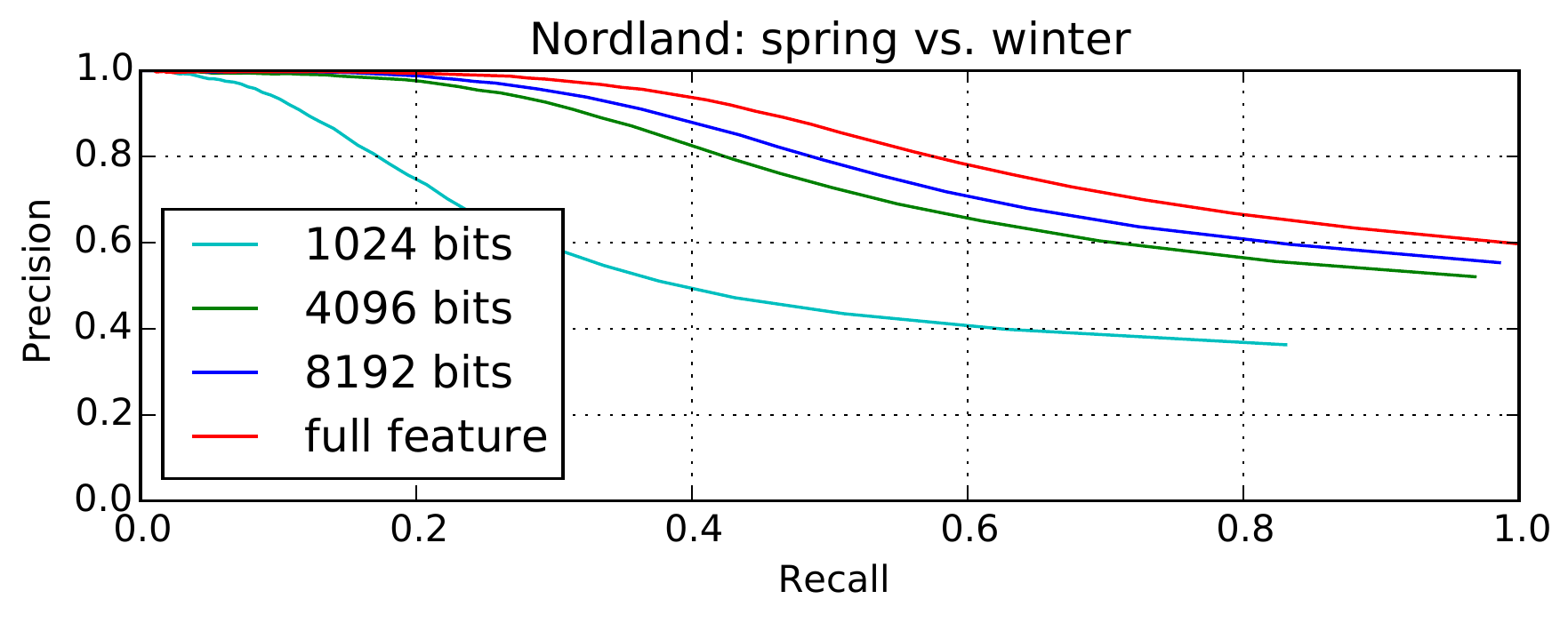}
    \includegraphics[width=0.9\linewidth]{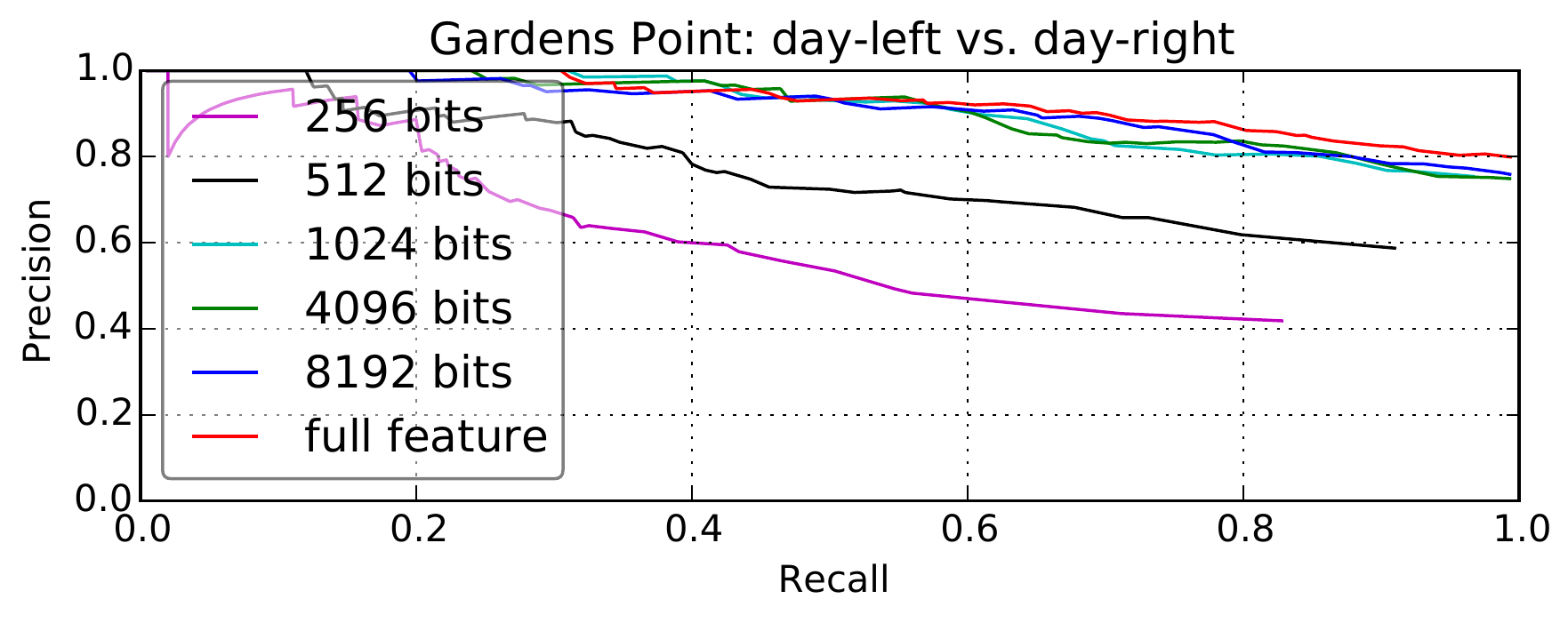}
   \end{center}
   \caption{The cosine distance over the full feature vector of 64896 floating
   point elements (red), can be closely approximated by the Hamming distance over bit
   vectors of length $2^{13}$ (blue) without losing much performance. This corresponds to a
   compression of 99.6\%. The bit vectors are calculated using the cosine
   similarity preserving Locality Sensitive Hashing method as proposed in \cite{Ravichandran05}.
   }
  \label{fig:gardensPoint:bits}
\end{figure}

Computing the cosine distance between many 64,896 dimensional \texttt{conv3}
feature vectors is an expensive operation 
and is a bottleneck of the ConvNet-based place recognition.
To speed this process up significantly, we propose to use a specialized variant of binary
locality-sensitive hashing that preserves the cosine similarity \cite{Charikar02}.
This hashing method leverages the property that the probability of a random
hyperplane separating two vectors is directly proportional to the angle between
these vectors. \cite{Ravichandran05} demonstrates how the cosine distance between two
high-dimensional vectors can be closely approximated by the Hamming distance between the
respective hashed bit vectors. The more bits the hash
contains, the better the approximation. 
We implemented this method and compare the place recognition performance
achieved with the hashed \texttt{conv3} feature vectors of
different lengths ($2^8 \dots 2^{13}$ bits) on the Nordland and Gardens Point
datasets in Fig. \ref{fig:gardensPoint:bits}. Using 8192 bits
retains approximately 95\% of the place recognition performance.
Hashing the original 64,896 dimensional vectors into 8192 bits corresponds to a data compression
of 99.6\%. Since the Hamming distance over bit vectors is a computationally
cheap operation, the best matching image among 10,000 candidates can be found
within 13.4 ms on a standard desktop machine. This corresponds to a speed-up
factor of 266 compared to using the cosine distance over the original
\texttt{conv3} features which required 3570 ms per 10,000 candidates.
Calculating the hashes requires 180 ms using a non-optimized Python
implementation. 
Table \ref{tbl:timing} summarizes the required time for the main
algorithmic steps. We can see that the hashing enables real-time place
recognition using ConvNet features on large scale maps of 100,000 places or more. 
\begin{table}[b]
  \begin{center}
    \begin{tabular}{@{}llll@{}}	
      \toprule
      & {\bf original \texttt{conv3}} & {\bf 4096 bits} & {\bf 8192 bits} \\ \midrule
      ConvNet feature & 15 ms  & 15 ms & 15 ms   \\
      hashing &  ---  &  100 ms  &  180 ms  \\
      match 100k candidates & 35,700 ms  &  89 ms & 134 ms  \\ \midrule
      frame rate & 0.03 Hz & 4.9 Hz & 3.0 Hz \\ \bottomrule 
    \end{tabular}
  \end{center}
  \caption{Runtime comparison between important algorithmic steps for 
  the original features and two different hashes on a desktop machine with
  a Nvidia Quadro K4000 GPU.
  }
\label{tbl:timing}
\end{table}
\subsection{Search Space Partitioning using Semantic Categorization}
\begin{figure}[t]
  \begin{center}
    \includegraphics[width=0.9\linewidth]{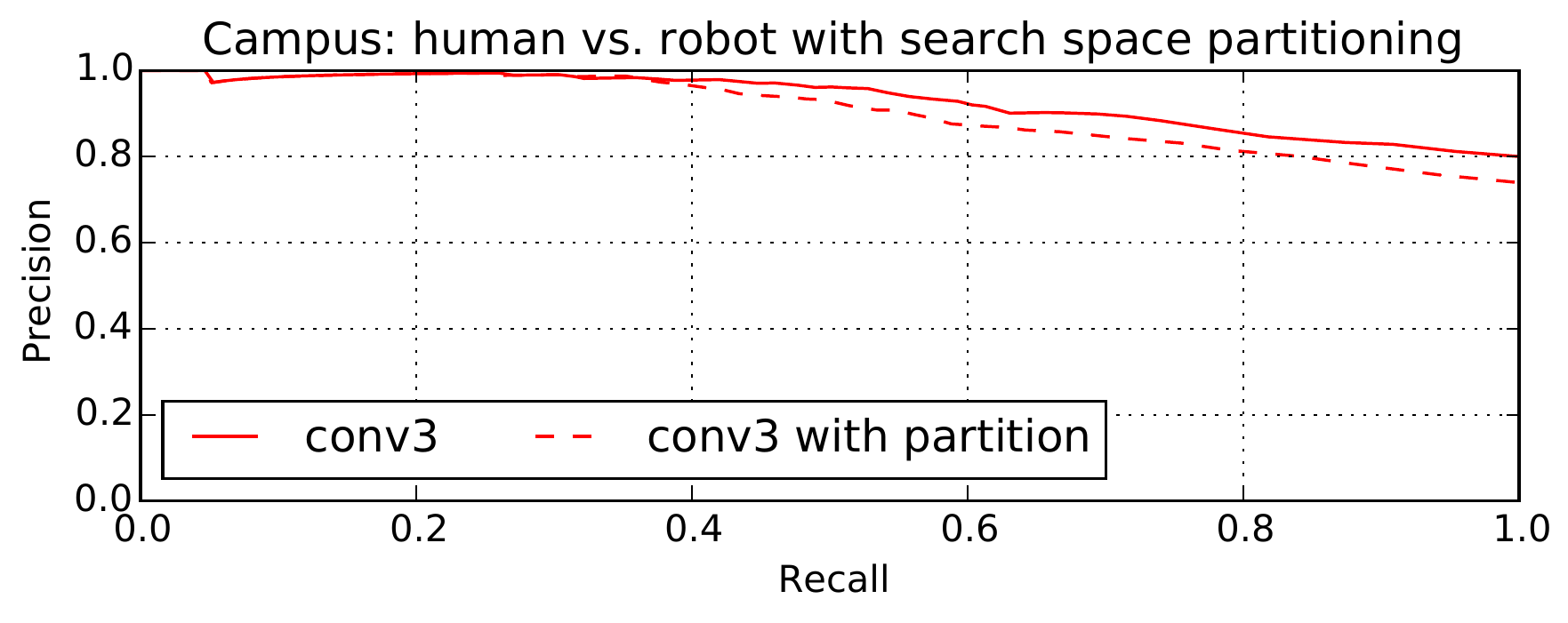}
   \end{center}
  \caption{Partitioning the search space decreases the performance slightly but
  can significantly reduce the time spent for the nearest neighbor search (76\%
  in this case).
  }
  \label{fig:campus:partition}
\end{figure}

We propose a novel approach to exploit the \emph{semantic} information encoded 
in the \emph{high-level} ConvNet features to partition the search space and constrain 
the nearest neighbor search to areas of similar semantic place categories such
as office, corridor, classroom, or restaurant. 
This can significantly shrink the search space in \emph{semantically versatile}
environments.
To discriminate between semantic place categories we train a nonlinear SVM
classifier using the \texttt{fc7} layer. The training has to be performed only
once on images from the SUN-397 database \cite{Xiao14}.
When performing place recognition, a separate index for each semantic class
$c_i$ is maintained that allows fast access to all stored \texttt{conv3}
features that were recorded at previously visited places with the probability of
belonging to $c_i$ above a threshold $\theta$. The nearest neighbor search can
then be limited to places that belong to the same semantic category as the
currently observed place.
We tested this approach on the Campus dataset and trained the classifier to
discriminate between 11 semantic classes. Partitioning the search space 
decreased the time spent for the nearest neighbor search by a factor of 4. 
Fig. \ref{fig:campus:partition} shows the place recognition performance
decreasing slightly when using this technique. However, there is an adjustable
trade-off between recognition performance and runtime requirements. Lowering the
threshold ($\theta=10\%$ was used here) allows more candidate matches to be
assessed, thus increasing recognition performance at the expense of the runtime.
\section{Comparing Different ConvNets for Place Recognition}
\begin{figure}[t]
  \begin{center}
    \includegraphics[width=0.9\linewidth]{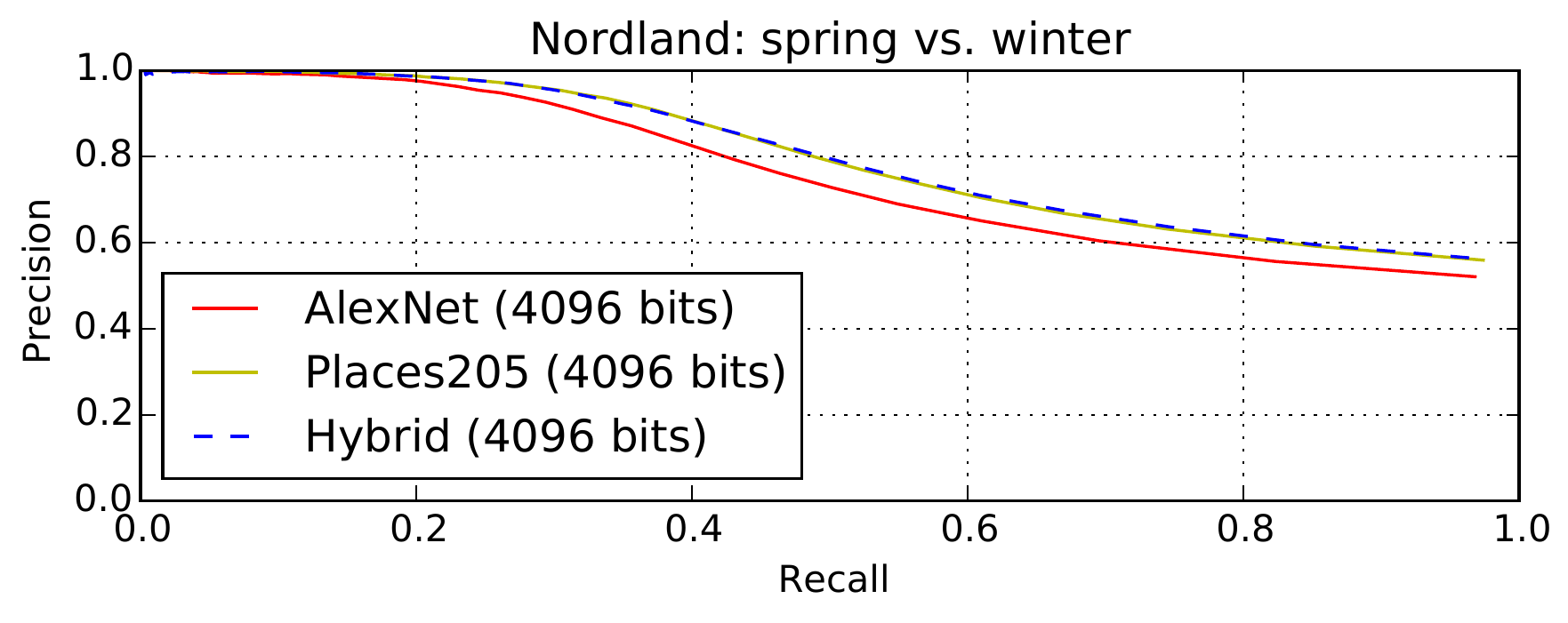}
   \end{center}
   \caption{The \texttt{Places205} and the \texttt{Hybrid} networks
   from \cite{Zhou14} perform slightly better at place recognition than
   \texttt{AlexNet} on the Nordland Spring vs. Winter dataset. The
   hashed \texttt{conv3} features were used for these experiments.}
  \label{fig:nordland:networks}
\end{figure}

All experiments described so far used the \texttt{AlexNet} ConvNet
architecture \cite{Krizhevsky12} as implemented by \texttt{Caffe} \cite{Jia14} that was
pre-trained for the task of object recognition on the ILSVRC dataset
\cite{Russakovsky14}. 
While preparing this paper, \cite{Zhou14} published the \texttt{Places205} 
and \texttt{Hybrid} networks. 
These two networks have the same principled architecture as \texttt{AlexNet} but
have been trained for scene categorization (\texttt{Places205}) or both tasks
(\texttt{Hybrid}).
We compare their place recognition performance with that of \texttt{AlexNet},
using the hashed \texttt{conv3} features.
Table \ref{tbl:networks} and Fig. \ref{fig:nordland:networks} summarize the results. 
Compared to \texttt{AlexNet}, \texttt{Places205} and the \texttt{Hybrid} network
perform slightly better under severe appearance changes. This 
could be explained by the fact that \texttt{AlexNet} is trained for \emph{object}
recognition while the two other networks are specialized for recognizing
\emph{scene} categories, i.e. they learned to discriminate places. However, in
the presence of viewpoint changes (Gardens Point left vs. right), the results
are inconclusive and \texttt{AlexNet} has a slight performance advantage. 
\begin{table*}[tb]
  \begin{center}
    \begin{tabular}{@{}lllllll@{}}	
      \toprule
      & && \multicolumn{4}{l}{\bf Place Recognition Performance (F-Score) using \texttt{conv3}}   \\ 
      & & & {\bf Nordland} &{\bf Gardens
      Point} \\
      & \multicolumn{2}{l}{\bf Pre-trained} & spring vs. & day-left vs. &
      day-right vs. & day-left vs.  \\
      {\bf Network}  & {\bf for Task} & {\bf on Dataset} &  winter &
      day-right & night-right & night-right \\      
      \midrule
      \texttt{AlexNet} & object recognition & ImageNet ILSVRC
      \cite{Russakovsky14} & 0.68  &  \bf 0.85  &  0.85 & \bf 0.48\\ 
      \texttt{Places205} & place categorization & Places Database \cite{Zhou14}
      & {\bf 0.71}  &   0.82  &  0.84  & 0.46 \\
      \texttt{Hybrid} & both of above tasks & Places + ILSVRC \cite{Zhou14} &
      {\bf 0.71} &  0.84  &  \bf 0.87 & 0.44 \\ \bottomrule
    \end{tabular}
  \end{center}
  \caption{The three convolutional networks we compare in this paper. All of them are
  implemented using \texttt{Caffe} \cite{Jia14}. Notice that none of them is
  specifically trained for the task of place recognition.}
\label{tbl:networks}
\end{table*}
\section{Conclusions}
Our paper presented a thorough investigation on the utility of ConvNet features
for the important task of visual place recognition in robotics. We presented a
novel method to combine the individual strengths of the high-level and mid-level
feature layers to partition the search space and recognize places under severe
appearance changes. We demonstrated for the first time that large-scale robust
place recognition using ConvNet features is possible when applying a specialized
binary hashing method. Our comprehensive study on four real world datasets
highlighted the individual strengths of mid- and high-level features with respect
to the biggest challenges in visual place recognition -- appearance and viewpoint
changes. A comparison of three state-of-the-art ConvNets revealed slight
performance advantages for the networks trained for semantic place categorization.
In subsequent work \cite{Suenderhauf15a} we applied the insights gained in this paper and extended the holistic approach presented here to a landmark-based scheme that addresses the remaining challenge of combined viewpoint \emph{and} appearance change robustness.
In future work we will investigate how training ConvNets specifically for the
task of place recognition under changing conditions can improve their
performance. 
\section*{Acknowledgements}
{\footnotesize 
\renewcommand{\baselinestretch}{0.2}
This research was conducted by the Australian Research Council Centre of Excellence for Robotic Vision (project number CE140100016). We want to thank Arren Glover and Will Maddern for collecting the Gardens Point and St. Lucia datasets.
}
\renewcommand{\baselinestretch}{1.0}
\bibliographystyle{IEEEtran}
\bibliography{bibfile}
\end{document}